\pdfoutput=1

\documentclass[11pt]{article}

\usepackage[]{emnlp2021}

\usepackage{times}
\usepackage{latexsym}

\usepackage[T1]{fontenc}

\usepackage[utf8]{inputenc}

\usepackage{microtype}

\usepackage{algorithm,algcompatible,amsmath}
\usepackage{graphicx}
\usepackage{url}
\usepackage{multirow}
\usepackage{comment}
\usepackage{amssymb}
\usepackage{pifont}
\usepackage{mathtools}
\usepackage{xcolor}
\usepackage{enumitem}
\usepackage{marvosym}
\usepackage{textcomp}
\usepackage{array}
\usepackage{booktabs}
\usepackage{multicol}
\usepackage{balance}
\usepackage{amsfonts}
\usepackage{amsmath}
\usepackage{xcolor}
\usepackage{calc}
\usepackage{amssymb}
\usepackage{diagbox}
\usepackage{algorithm}  
\usepackage{algorithmicx}  
\usepackage{algpseudocode} 

\makeatletter
\newcommand{\thickhline}{%
    \noalign {\ifnum 0=`}\fi \hrule height 1pt
    \futurelet \reserved@a \@xhline
}

\newcommand{\squishlist}{
\begin{list}{$\bullet$}
{ \usecounter{Lcount}
\setlength{\itemsep}{0pt}
\setlength{\parsep}{0pt}
\setlength{\topsep}{0pt}
\setlength{\partopsep}{0pt}
\setlength{\leftmargin}{2em}
\setlength{\labelwidth}{1.5em}
\setlength{\labelsep}{0.5em} } }
\newcommand{\squishend}{
\end{list} }

\newcommand{\modelname}{\texttt{MetaSRE}}

\newcommand{\f}{RCN}
\newcommand{\g}{RLGN}

\title{Semi-supervised Relation Extraction via Incremental Meta Self-Training}

\author{Xuming Hu$^1$, Chenwei Zhang$^{2\dagger}$, Fukun Ma$^1$, Chenyao Liu$^1$, Lijie Wen$^{1\dagger}$, Philip S. Yu$^{1,3}$\\
  $^1$Tsinghua University, $^2$Amazon, $^3$University of Illinois at Chicago\\
  $^1$\texttt{\{hxm19,mafk19,liucy19\}@mails.tsinghua.edu.cn}\\
  $^1$\texttt{wenlj@tsinghua.edu.cn}
  $^2$\texttt{cwzhang@amazon.com} 
  $^3$\texttt{psyu@uic.edu}\\
  }
\begin{document}
\maketitle

\begin{abstract}

To alleviate human efforts from obtaining large-scale annotations, Semi-Supervised Relation Extraction methods aim to leverage unlabeled data in addition to learning from limited samples.
Existing self-training methods suffer from the gradual drift problem, where noisy pseudo labels on unlabeled data are incorporated during training.
To alleviate the noise in pseudo labels, we propose a method called {\modelname}, where a Relation Label Generation Network generates quality assessment on pseudo labels by (meta) learning from the successful and failed attempts on Relation Classification Network as an additional meta-objective.
To reduce the influence of noisy pseudo labels, {\modelname} adopts a pseudo label selection and exploitation scheme which assesses pseudo label quality on unlabeled samples and only exploits high-quality pseudo labels in a self-training fashion to 
incrementally augment labeled samples for both robustness and accuracy. 
Experimental results on two public datasets demonstrate the effectiveness of the proposed approach. Source code is available\footnote{\url{https://github.com/THU-BPM/MetaSRE}\\\phantom{00} $^\dagger$Corresponding Authors.}.

\end{abstract}
\section{Introduction}\label{introduction}
Relation extraction plays a key role in transforming massive corpus into structured triplets (subject, relation, object). For example, given a sentence: \textit{The song was composed for a famous Brazilian musician}, we can extract a relation \textsc{Product-Producer} between two entities \textit{song} and \textit{musician}. These triples can be used in various downstream applications such as web search, sentiment analysis and question answering.
Current relation extraction methods ~\citep{zeng2016incorporating,zhang2017position} can discover the semantic relation that holds between two entities under supervised learning. However, these methods typically require lots of manually labeled data for model training. While in practice, these labeled data would be labor-intensive to obtain and error-prone due to human subjective judgments. 

A lot of work is being explored to alleviate the human supervision in relation extraction. Distant Supervision methods ~\citep{mintz2009distant,zeng2015distant}
leverage external knowledge bases to obtain annotated triplets as the supervision.
These methods make a strong assumption that the relation between entity pairs should not depend on the context, which leads to context-agnostic label noises and sparse matching results.
Semi-Supervised Relation Extraction method aims to leverage large amounts of unlabeled data to augment limited labeled data. There are two major ways to make fully use of unlabeled data, i.e., \textit{self-ensembling} method and \textit{self-training} method.
\begin{figure}[bp!]
    \centering
    \includegraphics[width=\linewidth]{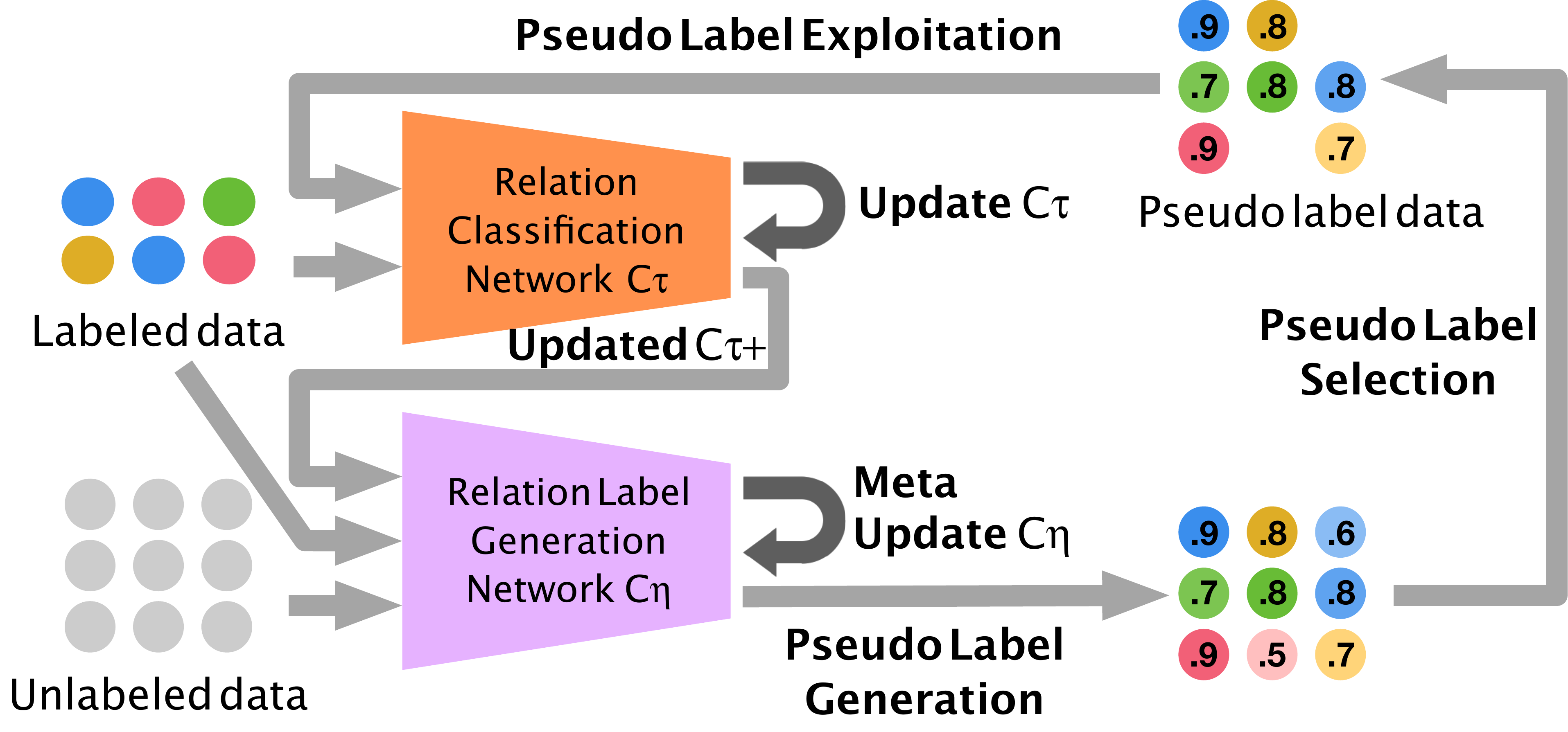}
    \caption{Semi-supervised Relation Extraction via Incremental Meta Self-Training. Relation Classification Network (RCN) uses both labeled data and unlabeled data with pseudo labels generated by the Relation Label Generation Network (RLGN). RLGN will be meta optimized by RCN using the labeled data and leverages pseudo label selection/exploitation scheme to obtain high-quality pseudo labels for RCN.}
    \label{fig:overview}
\end{figure}
Self-ensembling methods \citep{miyato2018virtual} assume that predictions on the unlabeled data by the model should remain unchanged, even if there are perturbations in the model parameters or training data. Self-ensembling methods usually suffer from insufficient supervision -- when labeled data is limited, the model is reluctant to acquire new relation knowledge that could be learned from the unlabeled data and thus impede further improvements.
On the contrary, self-training methods \citep{rosenberg2005semi} improve the predictive ability of the model by obtaining high-confidence labels from unlabeled data incrementally, and retraining the model on the updated labeled data. However, using self-training directly may introduce noisy pseudo labels inevitably, which hurts the model performance, known as the gradual drift problem \citep{liu2019self,li2020exploit}.

To alleviate the noise in pseudo labels, we design a method that can generate high-quality pseudo labels from the unlabeled dataset as additional supervision. We propose a semi-supervised learning framework which adopts meta-learning during pseudo label generation and automatically learns to reduce the influence of noisy pseudo labels. We also adopt a pseudo label selection and exploitation scheme during each self-training step to take full advantage of high-quality pseudo labels. As shown in Figure \ref{fig:overview}, the proposed framework has two networks, Relation Classification Network ({\f}) and Relation Label Generation Network ({\g}). {\f} encodes entity pair representations based on the context in which they are mentioned, and train a classifier with both labeled data and pseudo-labeled data generated by {\g} to classify the representations into relation categories. {\g} leverages pseudo label selection/exploitation scheme to calibrate confidence scores and obtain high-quality pseudo labels from unlabeled data for {\f}. {\g} is meta optimized by {\f} using the labeled data.

The main contributions of this work are as follows:
\begin{itemize}
    \item We propose a novel semi-supervised relation extraction framework {\modelname}, which adopts a meta learner network to prevent the model from drifting due to label noise and enables robust iterative self-training.
    \item We develop a label selection and exploitation scheme that explores all unlabeled samples and exploits high-confidence ones to get pseudo labels, 
    for effective and efficient self-training.
    \item We show that the semi-supervised model outperforms strong baselines and extensive experiments validate the effectiveness of the proposed model.
\end{itemize}

\section{Proposed Model}
The proposed Incremental Meta Self-training Relation Extraction ({\modelname}) consists of two networks: Relation Classification Network ({\f}) and Relation Label Generation Network ({\g}). As shown in Figure \ref{fig:overview}, the input of the {\f} is labeled data, which consists of sentences and relation mentions: ${[Sentence, Entity_{1}, Entity_{2}, Relation]}$. The goal of the network is to conduct relation classification using both labeled data and unlabeled ones with high-quality pseudo labels generated by the {\g}. In order to alleviate gradual drifts in traditional self-training scheme, the {\g} will be meta-optimized by {\f} using the labeled data to insure pseudo label quality. {\g} leverages pseudo label selection scheme to obtain high-confidence pseudo labels with less noises. {\f} exploits these pseudo labels in an iterative fashion and further improves its classification performance.

\subsection{Relation Classification Network ({\f})} 
The main purpose of {\f} is to extract the relational representation of context from sentences, and classify these features to get the corresponding relations. 
We assume that the named entities in the sentences have been recognized and marked in advance, and we need to focus on the binary relations which involve two entities.

\noindent\textbf{Contextualized Relation Encoder}

Since the relation between two entities in a sentence is often contained in the context of the two entities, the relation is carried through the contexts in which entities are expressed. In this work, we use pretrained deep bi-directional transformers networks: BERT \citep{devlin2019bert} to extract contextualized entity features.

For an input sentence containing a relation mention, two entities ${Entity_{1}}$ and ${Entity_{2}}$ are marked in advance. We follow the labeling mechanism adopted by \citet{zhang2015relation,soares2019matching} to enhance the position information of entities. For each sentence ${X = [x_{1}, .., x_{T}]}$, four reserved tokens ${[E1_{start}]}$, ${[E1_{end}]}$, ${[E2_{start}]}$, ${[E2_{end}]}$ are inserted to indicate entities' start and end positions:
\begin{equation}
\begin{split}
{X}=&\big[x_{1},...,[E1_{start}],x_{i},...,x_{j-1},[E1_{end}],\\
\qquad...,&[E2_{start}],x_{k},...,x_{l-1},[E2_{end}],...,x_{T}\big].
\end{split}
\end{equation}
The updated sentence is used as the input of the Contextualized Relation Encoder.
Instead of using the sentence-level output $[CLS]$ from BERT, we use the entity-level output $[E1_{start}]$, $[E2_{start}]$ and concatenate them and obtain a fixed-length vector $\mathbf{h}\in\mathbb{R}^{2\cdot{h_{R}}}$ as the contextualized entity pair representation:
\begin{equation}
\mathbf{h} = [\mathbf{h}_{[E1_{start}]}, \mathbf{h}_{[E2_{start}]}].
\end{equation}

\noindent\textbf{Relation Classification}\label{rc} 
Once we obtain contextualized entity pair representations, the contextualized relation encoder 
is used to classify these representations. 
When the labeled data that we used to train the classifier is insufficient, the classifier may have limited generalization ability.
High-quality pseudo labels generated on unlabeled data can help improve the classification performance. However, not all unlabeled data are helpful. 
In our model, in order to mitigate such noises, we construct the {\g} specifically to associate unlabeled sentences with high-quality pseudo labels. This network will be introduced in detail in Section \ref{generation network}.

{\f} learns to predict the right golden labels and pseudo labels. More specifically, we have:
\begin{equation}
    \mathbf{l}_n = C_{\tau}(X_{n, E1, E2}),
\end{equation}
where $\mathbf{l}_n$ is a probability distribution over the number of relations. $C_{\tau}$ consists of the Contextualized Relation Encoder module which converts ${X_{n, E1, E2}}$ into ${\mathbf{h} = [\mathbf{h}_{[E1_{start}]}, \mathbf{h}_{[E2_{start}]}]}$ and a fully connected dense layer that uses ${\mathbf{h}}$ for classification. There are a total of $N$ golden labels and $M$ pseudo labels: $S_{all} = \{g_1, g_2, ..., g_N, p_1, p_2, ..., p_M\}$. To distinguish pseudo labels from golden labels when optimizing $\tau$, we adopt the following classification loss:
\begin{equation}
\begin{split}\label{eq:f_loss}
\mathcal{L}_{{C}_\tau} = \sum_{n=1}^{N}\mathit{loss}(\mathbf{l}_n,\operatorname{one\_hot}(g_{n}))+\\
\sum_{m=1}^{M}{w_m}\cdot\mathit{loss}(\mathbf{l}_m,\operatorname{one\_hot}(p_{m})),
\end{split}
\end{equation}
where $loss$ is the cross entropy loss function. $\mathbf{l}_{n}$ is the inferred probability distribution of the labeled data. $\mathbf{l}_{m}$ is the inferred probability distribution of the unlabeled data via pseudo labeling. $\operatorname{one\_hot}(\cdot)$ returns an one-hot vector indicating the label assignment. Since we wish the module to confide less in pseudo labels than golden labels, therefore when optimizing parameter $\tau$ in classifier $C_{\tau}$, the loss terms on pseudo labels are used with a confidence coefficient $0\leq{w_m}\leq1$, which is generated by the {\g}. Note that in the first iteration, the second term for pseudo labels in Eq. \ref{eq:f_loss} is not applicable until subsequent iterations where pseudo labels are generated.

\subsection{Relation Label Generation Network ({\g})}\label{generation network} 
We aim to make full use of unlabeled data to improve the classification quality of {\f}. In this section, {\g} is introduced to generate pseudo labels for unlabeled data with a simple yet effective meta-update scheme.

Unlike the traditional self-supervised scheme, {\modelname} does not directly use the {\f} to classify unlabeled sentences with pseudo labels. 
As illustrated in Figure \ref{fig:overview}, we construct a {\g} to generate pseudo labels, which has the same architecture as the {\f} but is trained separately as a meta learner to label the unlabeled sentences and learn more about the distribution of relation mentions on pseudo labels through the {\f}.

The reason why we discard the classification network and retrain another network is to prevent the noise contained in the generated pseudo labels which would lead the sentence feature distribution to drift gradually \citep{zhang2016understanding,liu2019self}. For example, when the classifier incorrectly gives the unlabeled sentence: \textit{The \textbf{song} was composed for a famous Brazilian \textbf{musician}} a false pseudo label: \textsc{Content-Container} instead of \textsc{Product-Producer}, this falsely-labeled sentence will be added into pseudo label mentions, which will accumulate errors in the subsequent training.

To address this issue, we adopt a simple yet effective schema: we let the {\g} learn to effectively assess the quality of pseudo labels 
by (meta) learning from the successful and failed attempts using the most updated {\f} as an additional meta-objective. 
This schema can be seen as a form of meta-learning. In other words, the meta objective of the {\g} is to perform a derivative over the parameters on {\f}. In order to prevent noises in the generated pseudo labels from contaminating the overall objective, {\g} is tuned to generate pseudo labels using only the labeled data. These two networks have the same network structure but are initialized separately and trained completely independently. To distinguish, we denote the parameters of the {\g} as ${\eta}$. The meta objective is defined as follows:
\begin{equation}
\underset{\eta}{\operatorname{argmin}}\ \mathit{loss}\big(C_{\tau^{+}}(X_{n, E1, E2}),\operatorname{one\_hot}(g_{n})\big),
\end{equation}
where ${\tau^{+}}$ represents the parameters of the {\f} after one gradient update using the loss in Equation \ref{eq:f_loss}, $n$ comes from $N$ golden labels: $S_{G} = \{g_1, g_2, ..., g_N\}$:
\begin{equation}\label{tau+}
\tau^{+}\leftarrow\tau-\alpha\nabla_{\tau} \mathcal{L}_{C_{\tau}},
\end{equation}
where $\alpha$ is the learning rate. The trick in meta objective is to use the updated parameters ${\tau^{+}}$ to calculate the derivative and update ${\eta}$. This can make the ${\eta}$ learn more about the procedure of ${\tau}$ to learn pseudo labels. This trick was also adopted in other meta-learning frameworks such as \citet{finn2017model},  \citet{zhang2018fine} and \citet{liu2019self}. 

Therefore, the final optimization function is:

\begin{equation}
\begin{split}\label{eq.g_loss}
\mathcal{L}_{C_{\eta}}= \underset{\eta}{\operatorname{min}}\sum_{n=1}^{N}\big(\mathit{loss}((C_{\tau^{+}}(X_{n, E1, E2}),\\\operatorname{one\_hot}(g_{n}))\big)  
\end{split}
\end{equation}

After we optimize {\g} after an epoch, there are broadly two methods to make fully use of unlabeled data, i.e., \textit{self-ensembling} method and \textit{self-training} method. As we illustrate in Section \ref{introduction}, we adopt the self-training method and use {\g} to incrementally augment labeled data via iterative pseudo label generation.

\subsection{Pseudo label Selection and Exploitation}

\noindent\textbf{Selection}

In the selection stage,  {\g} selects the potential pseudo label for each unlabeled sample. We denote $m'$ for each unlabeled sample and $X_{m'}$ is one unlabeled data we use to obtain the pseudo label for. We consider the relation that corresponds to the maximum probability after softmax $\operatorname{argmax}(C_{\eta}(X_{m', E1, E2}))$ as the pseudo label and use the corresponding probability as the confidence score.
Since not all generated pseudo labels are equally informative, we sort all pseudo labels according to the confidence score in a descending order and
select the top $Z\%$ high-confidence pseudo labels as the final $M$ pseudo labels, denoted as $S_{P} = \{p_1, p_2, ..., p_M\}$.

\noindent\textbf{Exploitation}

When we complete pseudo label selection and obtain high-confidence pseudo labels, we need to exploit these pseudo labels. Although the obtained pseudo labels are of high confidence, simply treating them as golden labels may introduce noises and deteriorate the {\f} robustness. When adding pseudo labels to the labeled data $S_{G}$ which will be used to optimize the {\f} parameters, we use the maximum probability value from the output of {\g} as the weight:
\begin{equation}
w_{m}=\underset{m}{\operatorname{max}}(C_{\eta}(X_{m, E1, E2})),
\end{equation}
where $m$ comes from the final $M$ exploited pseudo labels. $w_m$ is not updated during the optimization of {\f}. 

\noindent\textbf{Incremental Self-Training}

Pseudo labeling all unlabeled data at once using limited labeled data is not ideal. In this work, we explore an incremental way to select high-quality pseudo labels and utilize them in batches to let the {\f} gradually improve as we obtain more high-quality pseudo labels from the {\g}. The whole Incremental Self-Training workflow is illustrated in Algorithm \ref{algorithm}.

\begin{algorithm}
\caption{Incremental Self-Training in {\modelname}}
\label{algorithm}
\begin{algorithmic}[1]
\Require Labeled data and unlabeled data. Unlabeled data are divided into 10 batches.
\State Train the Classification Network $C_{\tau}$ using labeled data (Eq. \ref{eq:f_loss}).
\For {each batch of 10\% unlabeled data}
\State Meta update the Generation Network $C_{\eta}$ using labeled data and the updated Classification Network $C_{\tau^{+}}$ (Eq. \ref{eq.g_loss}).
\State Generate pseudo labels using the Generation Network $C_{\eta}$ for unlabeled data in this batch.
\State Use Pseudo Label Selection to select top $Z\%$ high-confidence pseudo labels.
\State Update the Classification Network $C_{\tau}$ using labeled data and high-confidence pseudo labels (Eq. \ref{eq:f_loss} \& \ref{tau+}).
\EndFor
\end{algorithmic} 
\end{algorithm}

\section{Experiments}
We first introduce datasets, experimental settings and evaluation metrics, and then present the performance comparison with baseline models. A detailed analysis is presented to show the advantages of each module.
\subsection{Datasets}
We use two public relation extraction datasets with different characteristics in our experiments:
(1) \textbf{SemEval} \citep{hendrickx2010semeval}: SemEval 2010 Task 8 provides a standard benchmark dataset and is widely used in the evaluation of relation extraction models. Its training, validation, test set contain 7199, 800, 1864 relation mentions respectively, with 19 relation types in total (including \textit{no\_relation}), of which \textit{no\_relation} percentage is 17.4\%.
(2) \textbf{TACRED} \citep{zhang2017position}: TAC Relation Extraction Dataset is a large-scale crowd-sourced relation extraction dataset. The corpus is collected from all the prior TAC KBP shared tasks, and follows TAC KBP relation scheme. Its training, validation, test set contain 75049, 25763, 18659 relation mentions respectively, with 42 relation types in total (including \textit{no\_relation}), of which \textit{no\_relation} percentage is 78.7\%.

For the two datasets above, their named entities in the sentences have been recognized and marked in advance. In terms of data characteristics, TACRED is far more complicated than SemEval because it has more relation types and more skewed distribution on negative mentions.

\begin{table*}[t]
\centering
  \resizebox{\linewidth}{!}{%
\begin{tabular}{l|ccc|ccc|ccc}
\thickhline
\multirow{2}{*}{\begin{tabular}[c]{@{}l@{}}\diagbox{Methods}{Labeled Data}\end{tabular}} & \multicolumn{3}{c}{5\%}            & \multicolumn{3}{|c}{10\%}            & \multicolumn{3}{|c}{30\%}            \\ \cline{2-10} 
& Precision & Recall & F1             & Precision & Recall & F1             & Precision & Recall & F1             \\ \hline
 LSTM                                                            & 25.34 ± 2.17     & 20.68 ± 4.01  & 22.65 ± 3.35          & 36.93 ± 6.23     & 29.65 ± 7.13  & 32.87 ± 6.79          & 64.80 ± 0.67     & 62.98 ± 0.66  & 63.87 ± 0.65          \\ 
PCNN                                                               & 42.87 ± 4.56    & 40.71 ± 4.36  & 41.82 ± 4.48          & 53.67 ± 1.52     & 49.23 ± 2.34  & 51.34 ± 1.87          & 64.52 ± 0.59     & 62.87 ± 0.53  & 63.72 ± 0.51        \\ 
PRNN          & 56.13 ± 1.29     & 54.52 ± 1.43  & 55.34 ± 1.08          & 61.70 ± 1.16     & 63.61 ± 2.07  & 62.63 ± 1.42          & 69.66 ± 2.12     & 68.72 ± 2.57  & 69.02 ± 1.01      \\
BERT              & 73.12 ± 1.34      & 72.21 ± 1.21  & 72.71 ± 1.24          & 74.71 ± 1.02     & 72.81 ± 0.98  & 73.93 ± 0.99          & 79.10 ± 0.89     & 81.05 ± 0.82  & 80.55 ± 0.87      \\\hline \hline 
 Mean-Teacher$_{BERT}$                                                             & 70.33 ± 3.45     & 68.55 ± 4.23  & 69.05 ± 3.89          & 74.01 ± 1.89     & 72.08 ± 1.32  & 73.37 ± 1.42          & 79.09 ± 0.89     & 82.23 ± 0.78  & 80.61 ± 0.81          \\ 
Self-Training$_{BERT}$                                                               & 73.10 ± 1.21     & 70.01 ± 1.83  & 71.34 ± 1.68          & 75.54 ± 1.07     & 73.00 ± 1.18  & 74.25 ± 1.10          & 80.92 ± 0.99    & 82.39 ± 0.67   & 81.71 ± 0.79        \\ 
DualRE$_{BERT}$                                                      & 73.32 ± 1.87     & 77.01 ± 1.67  & 74.35 ± 1.76          & 75.51 ± 1.21     & 78.81 ± 1.01  & 77.13 ± 1.10          & 81.30 ± 0.81      & 84.55 ± 0.52  & 82.88 ± 0.67      \\ 
MRefG$_{BERT}$                                                       & 73.04 ± 1.43     & 78.29 ± 1.21  & 75.48 ± 1.34          & 76.32 ± 1.00     & 79.76 ± 0.81  & 77.96 ± 0.90          & 81.75 ± 0.78      & 84.91 ± 0.63  & 83.24 ± 0.71   \\                        
{\modelname}$_{BERT}$                                        & \textbf{75.59 ± 0.92}     & \textbf{81.40 ± 0.91}  & \textbf{78.33 ± 0.92} & \textbf{78.05 ± 0.87}      & \textbf{82.29 ± 0.72}  & \textbf{80.09 ± 0.78} & \textbf{82.01 ± 0.36}     & \textbf{87.95 ± 0.56}   & \textbf{84.81 ± 0.44}     \\  \hline\hline
BERT w. gold labels                              & 82.43 ± 0.38     & 84.94 ± 0.21  & 83.64 ± 0.28          & 82.87 ± 0.36     & 86.16 ± 0.32  & 84.40 ± 0.34          & 86.28 ± 0.21     & 87.93 ± 0.27   & 87.08 ± 0.23    \\ \thickhline

\end{tabular}
}
\caption{Performance on SemEval with various amounts of labeled data and 50\% unlabeled data}\label{tab:SemEval}
\end{table*}
\begin{table*}[t]
\centering
  \resizebox{\linewidth}{!}{%
\begin{tabular}{l|ccc|ccc|ccc}
\thickhline
\multirow{2}{*}{\begin{tabular}[c]{@{}l@{}}\diagbox{Methods}{Labeled Data}\end{tabular}}  & \multicolumn{3}{c}{3\%}            & \multicolumn{3}{|c}{10\%}            & \multicolumn{3}{|c}{15\%}            \\ \cline{2-10} 
& Precision & Recall & F1             & Precision & Recall & F1             & Precision & Recall & F1             \\ \hline
LSTM                                                             & 40.63 ± 6.42     & 23.12 ± 4.58  & 28.68 ± 4.29          & 50.43 ± 0.97      & 43.18 ± 1.38  & 46.79 ± 0.99         & 55.69 ± 0.57     & 44.23 ± 0.62  & 49.42 ± 0.59          \\ 
PCNN                                                                 & 58.30 ± 7.43     & 37.78 ± 6.93  & 44.02 ± 5.23          & 64.64 ± 7.51     & 42.10 ± 4.94  & 50.35 ± 3.28          & 67.92 ± 0.51      & 42.09 ± 0.46  & 52.50 ± 0.39        \\ 
PRNN                                                        & 49.12 ± 5.23     & 33.21 ± 2.39  & 39.11 ± 1.92          & 53.71 ± 2.73      & 51.81 ± 1.72  & 52.23 ± 1.20          & 58.10 ± 2.78     & 51.05 ± 2.00  & 54.55 ± 1.92      \\
BERT                                                        & 49.82 ± 4.98     & 34.21 ± 3.19  & 41.11 ± 3.88          & 55.71 ± 2.17      & 52.81 ± 1.83  & 54.23 ± 1.67          & 59.10 ± 1.29     & 53.05 ± 1.02  & 56.55 ± 0.82      \\\hline \hline 
 Mean-Teacher$_{BERT}$                                                               & 50.04 ± 3.21     & \textbf{38.45 ± 2.73}  & 44.34 ± 1.78          & 56.02 ± 2.17     & 49.52 ± 1.09  & 53.08 ± 1.01          & 57.01 ± 1.02     & 51.24 ± 1.49   & 53.79 ± 1.38           \\ 
Self-Training$_{BERT}$                                                              & 48.01 ± 1.29     & 36.99 ± 1.56  & 42.11 ± 1.04          & 56.23 ± 0.67     & 52.02 ± 0.78  & 54.17 ± 0.53          & 59.25 ± 0.79     & 53.91 ± 0.33  & 56.52 ± 0.40          \\ 
DualRE$_{BERT}$                                                                     & 57.99 ± 1.77     & 35.21 ± 1.32   & 43.06 ± 1.73          & 60.00 ± 0.92     & 52.52 ± 0.37  & 56.03 ± 0.55          & 61.03 ± 0.54     & 55.24 ± 0.84  & 57.99 ± 0.67          \\ 
MRefG$_{BERT}$                                                      & 56.31 ± 1.72     & 36.25 ± 1.22  & 43.81 ± 1.44          & 59.25 ± 1.27     & 51.93 ± 1.88  & 55.42 ± 1.40          & 61.02 ± 0.82      & \textbf{55.61 ± 0.69}  & 58.21 ± 0.71   \\
{\modelname}$_{BERT}$    & \textbf{58.96 ± 1.00}     & 37.66 ± 1.26  & \textbf{46.16 ± 1.02} & \textbf{60.49 ± 0.92}     & \textbf{53.69 ± 0.58}  & \textbf{56.95 ± 0.34} & \textbf{65.03 ± 0.43}     & 54.02 ± 0.45  & \textbf{58.94 ± 0.36} \\ \hline \hline                        
BERT w. gold labels       & 65.92 ± 0.37     & 60.13 ± 0.48  & 62.93 ± 0.41         & 67.26 ± 0.29     & 60.42 ± 0.21  & 63.66 ± 0.23           & 68.01 ± 0.28     & 61.76 ± 0.26  & 64.69 ± 0.29    \\ \thickhline

\end{tabular}
}
\caption{Performance on TACRED with various amounts of labeled data and 50\% unlabeled data.}\label{tab:TACRED}
\end{table*}

\subsection{Baselines and Evaluation metrics}
As the proposed framework {\modelname} is general to integrate different contextualized relation encoders. We first compare several widely used supervised relation encoders such as \textbf{LSTM} \citep{hochreiter1997long}, \textbf{PCNN} \citep{zeng2015distant}, \textbf{PRNN} \citep{zhang2017position}, \textbf{BERT} \citep{devlin2019bert} and train them only on the labeled dataset, then adopt the best performing one as the base encoder for {\modelname} as well as for other representative semi-supervised approaches for a fair comparison.

We select four representative approaches from different categories of semi-supervised learning methods as our baselines:
\squishlist
 \item \textbf{Self-Training} \citep{rosenberg2005semi} uses a model to predict on unlabeled data recursively, and adds the pseudo labels to the labeled data. The model improves itself by retraining on the updated labeled dataset.
 \item \textbf{Mean-Teacher} \citep{tarvainen2017mean} encourages different variants of the model to make consistent predictions on similar inputs. The model is optimized by perturbation-based loss and training loss jointly.
 \item \textbf{DualRE} \citep{lin2019learning} leverages sentence retrieval as a dual task for relation extraction. The model combines the loss of a prediction module and a retrieval module to obtain the information corresponding to sentences and relations in unlabeled data.
 \item \textbf{MRefG} \citep{li2020exploit} is the state-of-the-art method that constructs reference graphs, including entity reference, verb reference, and semantics reference to semantically or lexically connect the unlabeled samples to the labeled ones.
\squishend

Finally, we present another model: \textbf{BERT w. gold labels} to demonstrate the upper bound of the performance for all the semi-supervised methods. This model trains the BERT base model using both the different proportion of labeled data and the gold labels of the unlabeled data, which means that all unlabeled data are correctly labeled.

For the evaluation metrics, we consider F1 score as the main metric while precision and recall serve as auxiliary metrics. Note that following \citet{li2020exploit}, the correct prediction of \textit{no\_relation} is ignored.

\subsection{Implementation Details}\label{implementation details} 
For all datasets, strictly following the settings used in \citet{li2020exploit}, we divide the training set into labeled and unlabeled sets of various sizes according to the stratified sampling which could ensure the distribution of the label will not change. We sampled 5\%, 10\%, and 30\% of the training set as labeled sets for the SemEval dataset, and the sampled 3\%, 10\%, and 15\% of the training set as labeled set for the TACRED dataset. For both datasets, we sampled 50\% of the training set as the unlabeled set. For all models that use unlabeled data in an incremental way, we follow the setting by \citet{li2020exploit} and fix the increased amount of unlabeled data as 10\% per iteration for a fair comparison, which means the unlabeled dataset will be exhausted after 10 iterations.

For the {\f}, in the Contextualized Relation Encoder module, we use the pretrained BERT-Base + Cased as the initial parameter. We use the BERT default tokenizer and set max-length as 128 to preprocess dataset. We use BertAdam with $1e{-4}$ learning rate to optimize the loss. In the Relation Classification module, we use a fully connected layer with the following dimensions: ${2\cdot{h_{R}}}$-${h_{R}}$-label\_size, where $h_R=768$. Learning rate is set as $1e{-4}$ and warmup to 0.1.

{\g} has the same structure and learning rate as the {\f}. We also use the pretrained BERT-Base + Cased as the initial parameter in the Contextualized Relation Encoder module. In the Pseudo label Selection module, each increment will generate 10\% of unlabeled data as pseudo labels. $Z\%=90\%$ is used.

\subsection{Main Results}
Table \ref{tab:SemEval} and \ref{tab:TACRED} show the experimental results on SemEval and TACRED dataset when adopting various labeled data and 50\% unlabeled data. We conduct 5 runs of training and testing then report the mean and standard deviation results. Considering that BERT \citep{devlin2019bert} has achieved the SOTA performance in all base encoders, we adopt BERT as the base encoder for all methods. We can also observe that MRefG achieves the best performance among all the baselines, which is considered as the previous SOTA method. The proposed model {\modelname} outperforms all baseline models consistently on F1. {\modelname} on average achieves 2.18\% higher F1 on SemEval dataset and 1.54\% higher F1 on TACRED dataset across various labeled data when comparing with MRefG. When considering standard deviation, {\modelname} performs more robust than all the baselines.

From Table \ref{tab:SemEval} and \ref{tab:TACRED}, when we fix unlabeled data to 50\% of the training set, we can find that with the increasing of labeled data, the performance of the model is effectively improved. In term of less labeled data, {\modelname} performs much better than the previous state-of-the-art model: MRefG. For example, {\modelname} achieves 2.85\% higher F1 performance with 5\% labeled data on SemEval and 2.35\% higher F1 performance with 3\% labeled data on TACRED, which indicate the effectiveness of our model under low-resource situations.

\begin{figure}[bt!]
    \centering
    \includegraphics[width=0.49\linewidth]{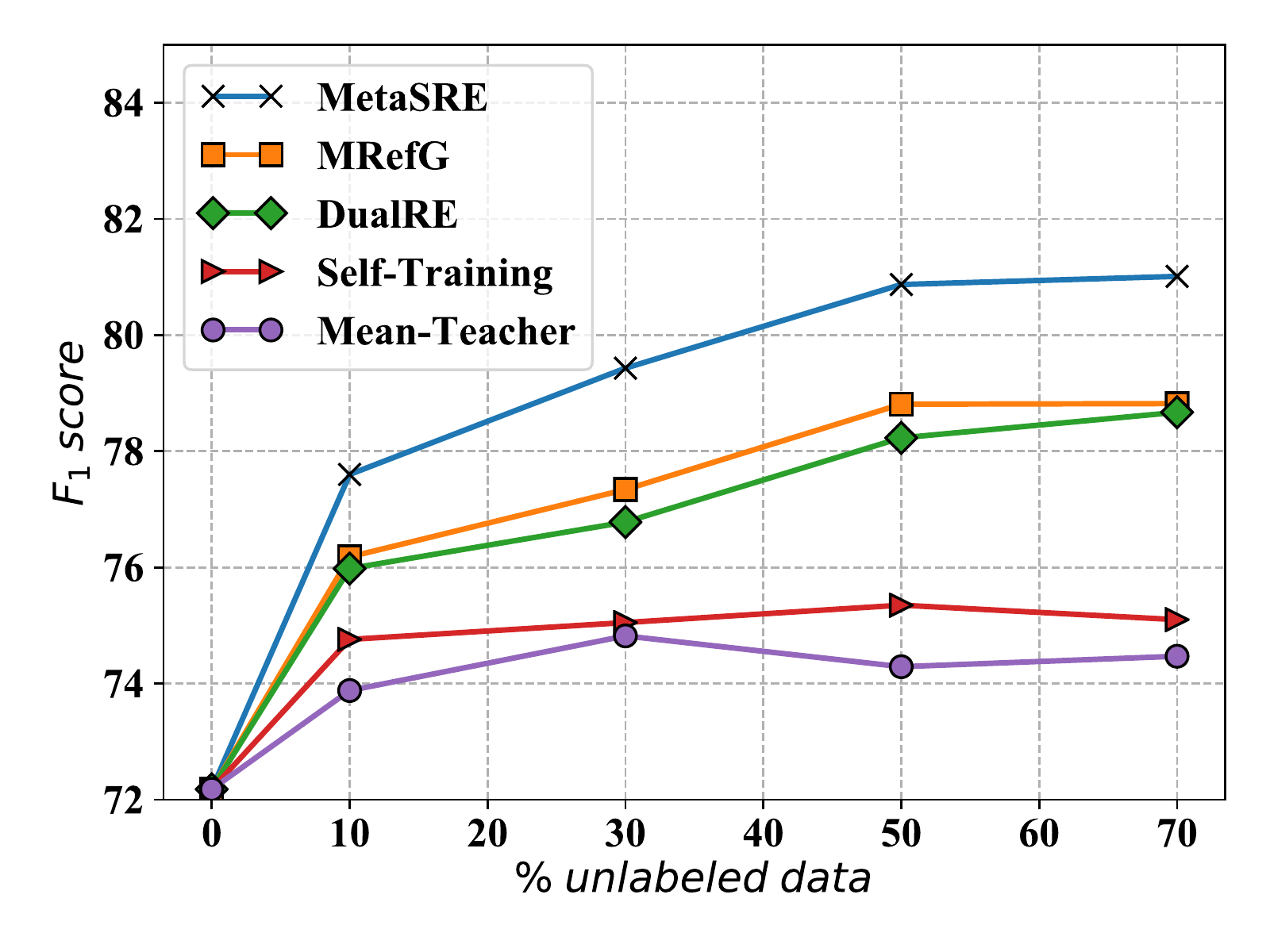}
    \includegraphics[width=0.49\linewidth]{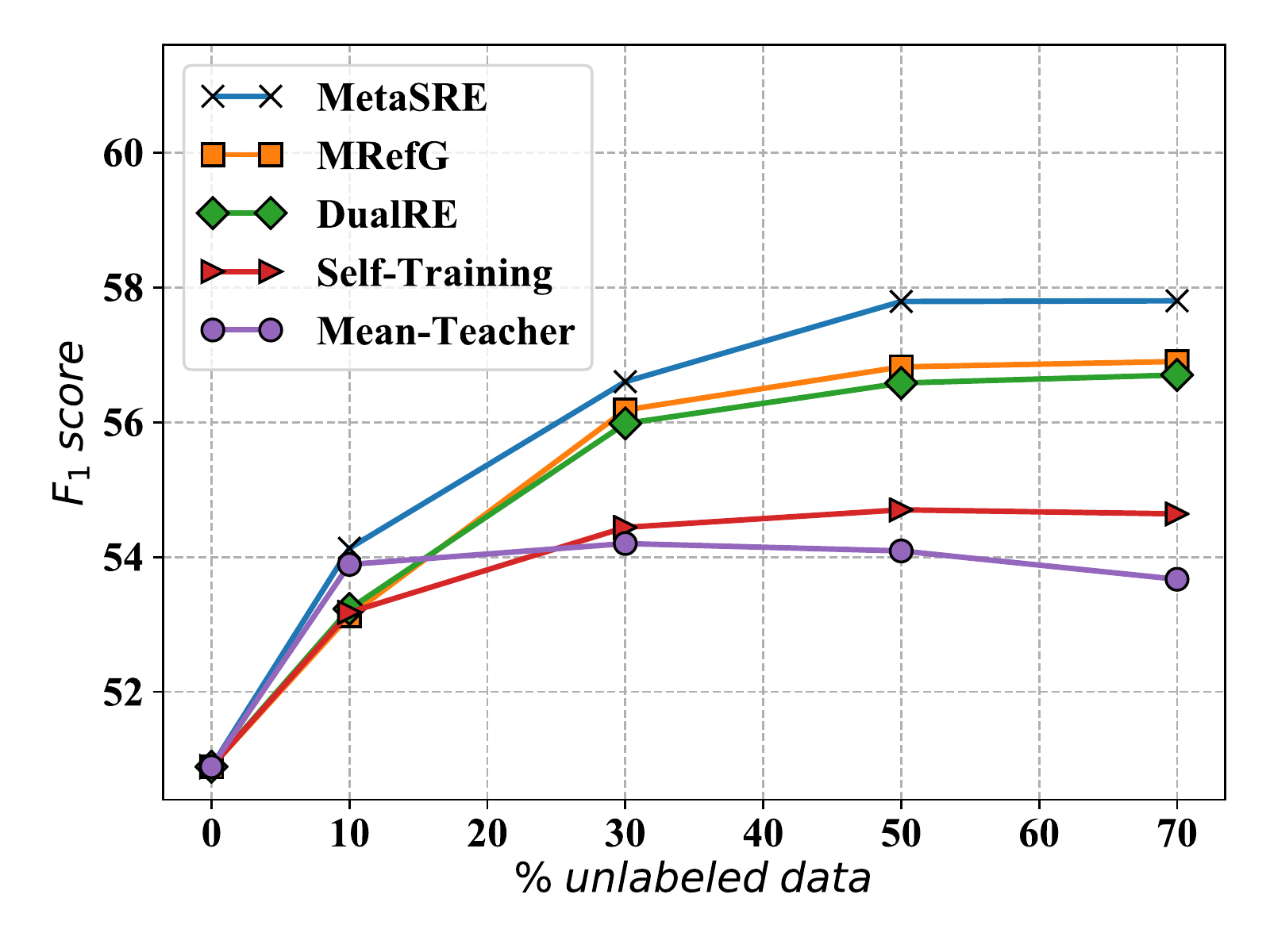}
    \caption{$F_{1}$ Performance with various unlabeled data and 10\% labeled data on SemEval (left) and TACRED (right).}
    \label{fig:unlabel}
\end{figure}

To investigate how models leverage different amounts of unlabeled data for performance improvement, we fix the amount of labeled data and compare the performance of the model with different amounts of unlabeled data. We report the F1 performance for SemEval and TACRED with 10\% labeled data and adopt unlabeled data as 10\%, 30\%, 50\%, 70\%. Since the labeled and unlabeled data are from the training set, we can provide up to 70\% unlabeled data. From Figure \ref{fig:unlabel}, we could see all semi-supervised methods have performance gains by using unlabeled data and {\modelname} achieves consistent and better F1 performance than other baselines under different ratios of unlabeled data. 
\begin{table}[]
\centering
\resizebox{\linewidth}{!}{
\begin{tabular}{lcccc}
\toprule
Model/$F_{1}$ Performance  & 5\%SemEval  & 30\%SemEval & 3\%TACRED & 15\%TACRED  \\
\midrule
{\modelname} & \textbf{78.33} & \textbf{84.81} & \textbf{46.16} & \textbf{58.94}  \\
w/o Meta Learning & 75.01 & 82.25 & 43.12 & 57.33  \\
w/o Pseudo Label Selection & 75.02 & 83.29 & 43.96 & 58.16  \\
w/o Pseudo Label Exploitation & 77.52 & 83.88 & 44.98 & 58.28  \\
\bottomrule
\end{tabular}}
\caption{Ablation study of {\modelname} on two datasets.}
\label{tab:ablation study}
\end{table}

\noindent\textbf{Ablation Study}

We conduct ablation study to show the effectiveness of different modules in {\modelname}. {\modelname} w/o Pseudo Label Selection is the proposed model without selecting Top ${Z\%}$ samples among all pseudo labels according to their confidences and adds all pseudo labels with weightings to the labeled data. {\modelname} w/o Pseudo Label Exploitation treats pseudo labels as having the same confidence score as golden labels, and does not distinguish between pseudo labels and golden labels when optimizing the {\g}. {\modelname} w/o Meta Learning uses the same {\f} for both label generation and classification. This is also equivalent to the Self-Training method with the addition of Pseudo Label Selection and Exploitation modules.

From ablation results in Table \ref{tab:ablation study}, we found that all modules contribute positively to the improved performance. More specifically, Meta Learning and Pseudo Label Selection modules do impact the performance: performances on {\modelname} w/o Meta Learning and {\modelname} w/o Pseudo Label Selection are deteriorated by 2.63\% and 1.88\% on F1 averaged over two datasets. Pseudo Label Exploitation gives 0.90\% performance boost in average over F1 on two datasets when comparing with the hard-weighting alternative.

\subsection{Analysis and Discussion}
\noindent\textbf{Effectiveness of Meta Learning}

The main purpose of the Meta Learning module is to generate pseudo labels with less noise and higher accuracy, which could prevent the {\f} from drifting gradually.
From the Ablation Study we can conclude the Meta Learning module can effectively improve the results. In order to explore how and why this module is effective, we use {\modelname} and {\modelname} w/o Meta Learning to explore the quality of the generated pseudo labels on the two datasets respectively. We sample the SemEval dataset with 30\% labeled data and the TACRED dataset with 15\% labeled data, and both with 50\% unlabeled data.

\begin{figure}[bt!]
    \centering
    \includegraphics[width=0.49\linewidth]{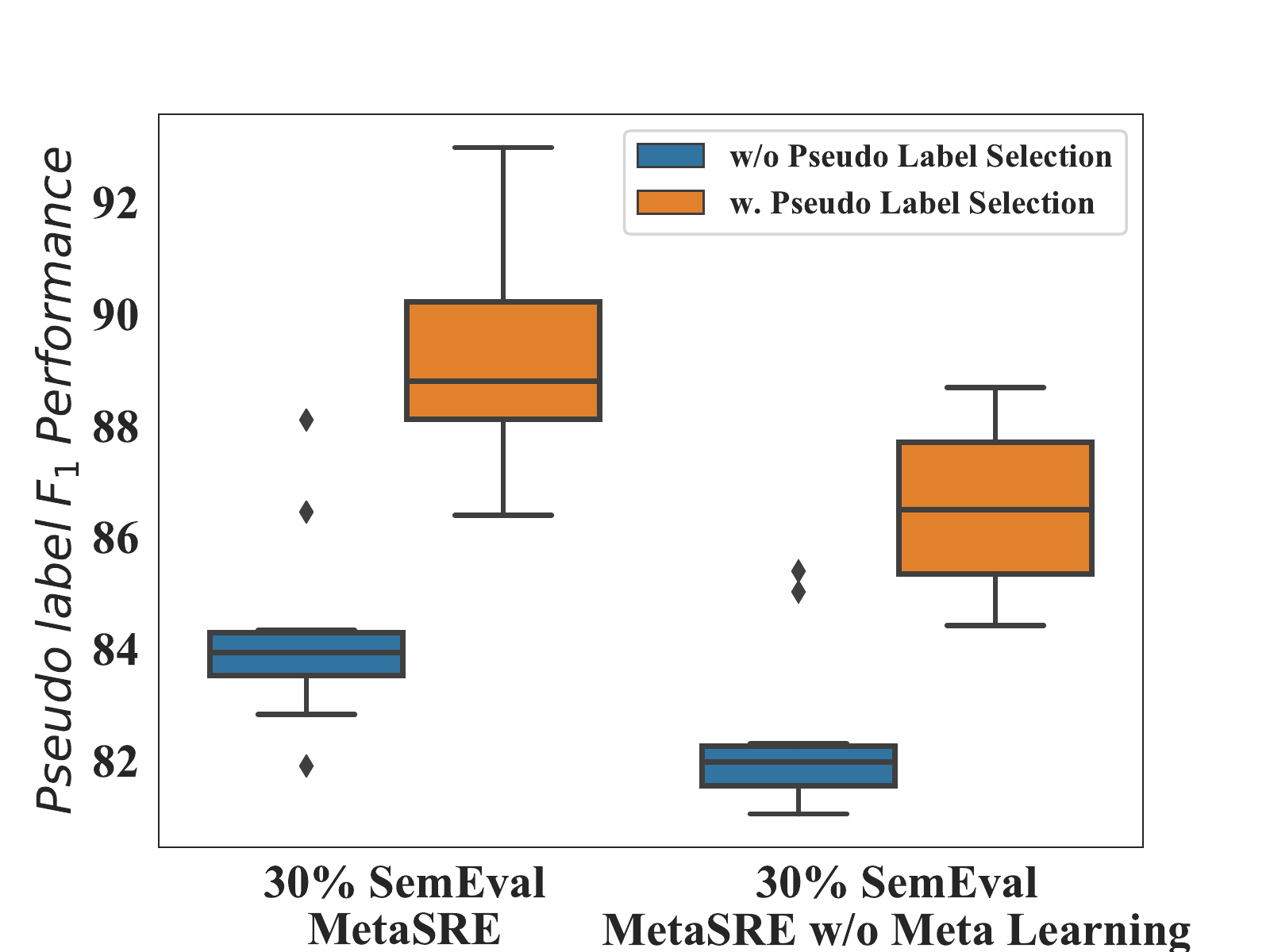} 
    \includegraphics[width=0.49\linewidth]{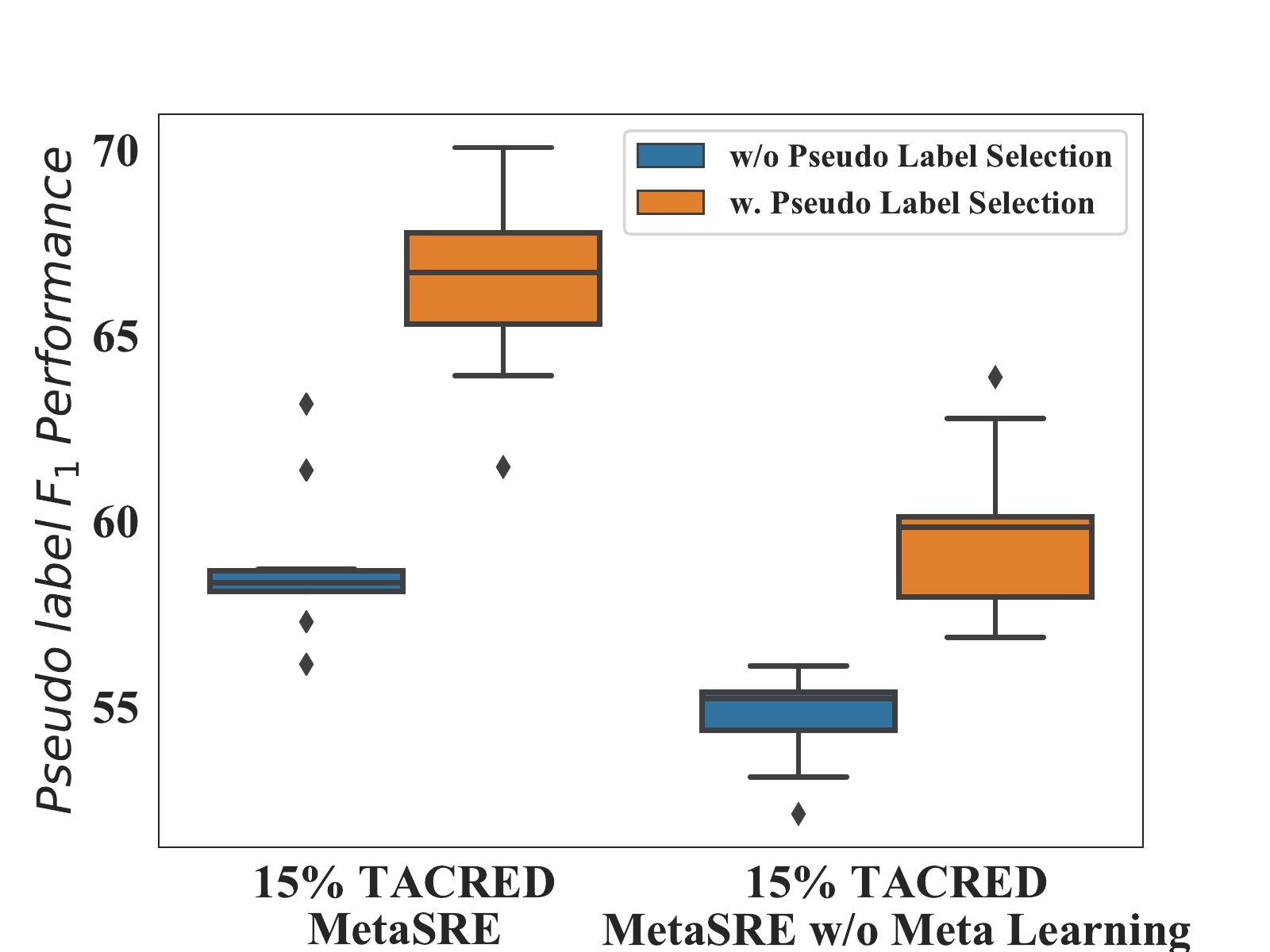}
    \caption{Pseudo label $F_{1}$ Performance with different modules based on SemEval (left) and TACRED (right).}
    \label{fig:meta-learning}
\vspace{-0.1in}
\end{figure}
\begin{figure}[bt!]
    \centering
    \includegraphics[width=0.49\linewidth]{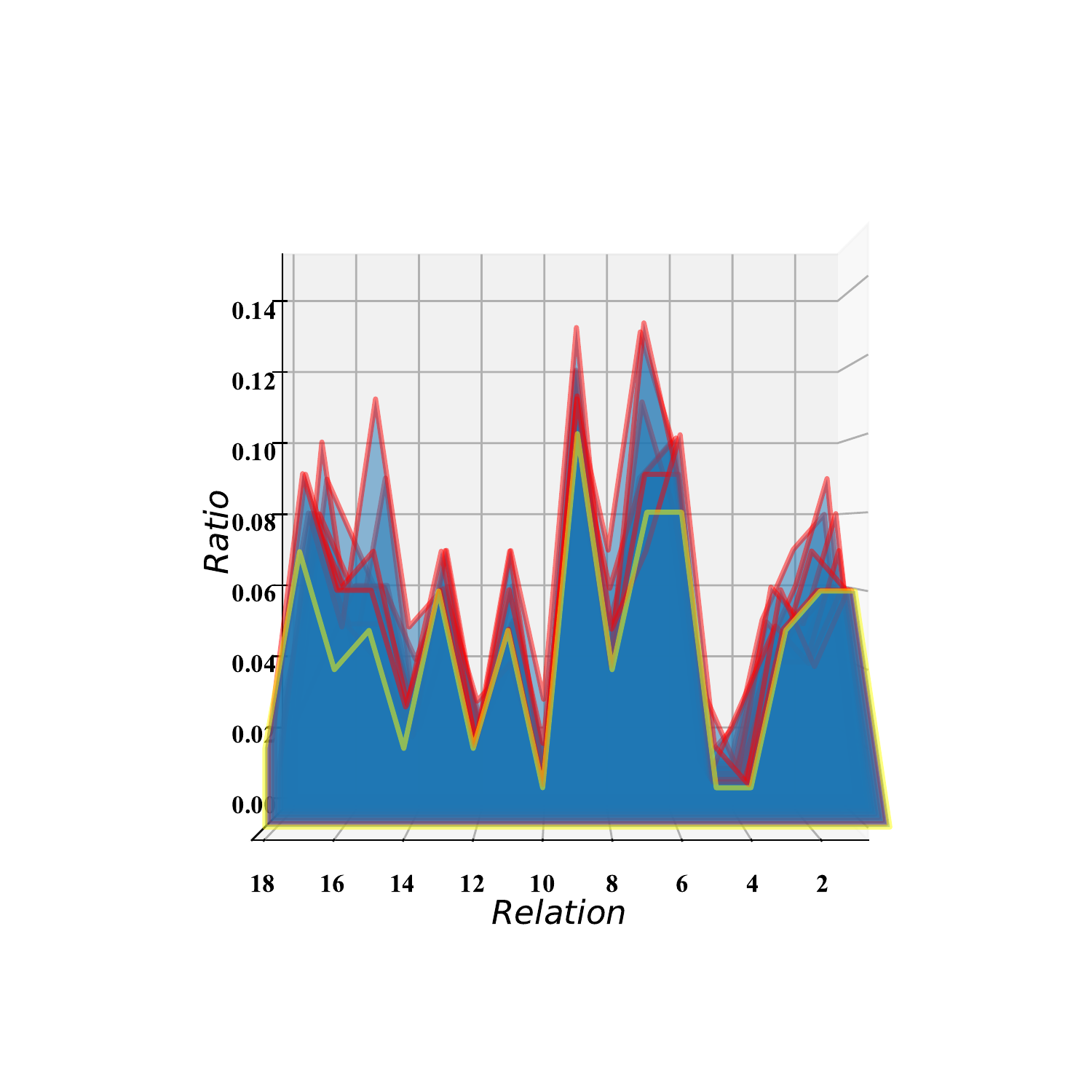} 
    \includegraphics[width=0.49\linewidth]{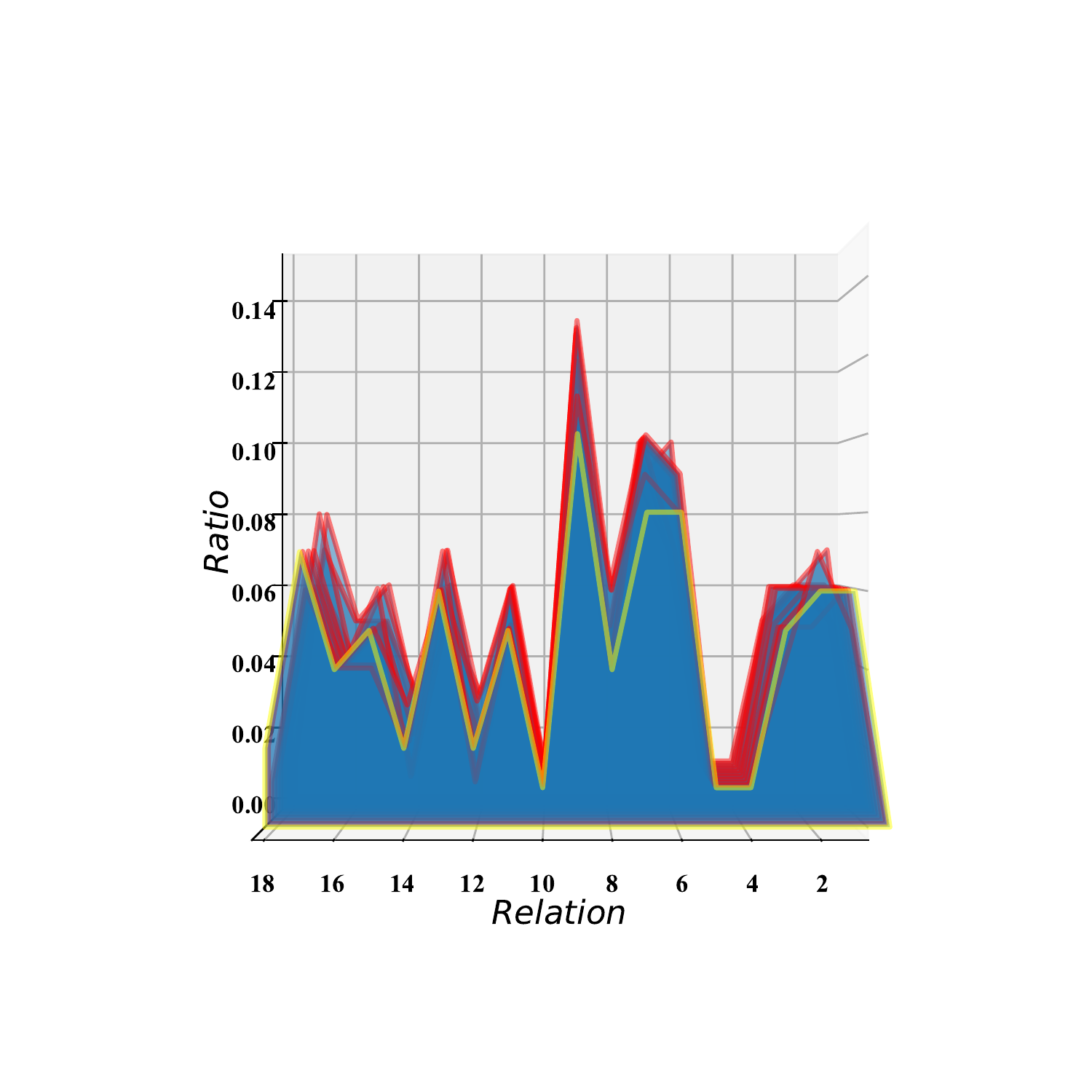}
    \caption{Pseudo label distributions generated by different iterations ({\color{red}red}) of {\modelname} w/o Meta Learning (left) and {\modelname} (right). {\color{yellow}Yellow} line is the gold label distribution.}
    \label{fig:label_distribution}
\vspace{-0.1in}
\end{figure}

From Figure \ref{fig:meta-learning}, we can find that:
(1) No matter whether it leverages Pseudo Label Selection or not, the Meta Learning module can continuously generate pseudo labels with higher quality. (2) After the Pseudo Label Selection, the quality of the pseudo labels generated by the Meta Learning module will be further improved. (3) On the TACRED dataset which has more skewed label distributions, Meta Learning module improves the quality of pseudo labels even more, which  means the Meta Learning module could help {\modelname} learn more accurate relational labels.

To validate meta learner network could prevent {\modelname} from drifting due to label noise and enables robust iterative self-training, in Figure \ref{fig:label_distribution}, we use yellow line to represent the gold label distributions and red lines to represent the pseudo label distributions which are generated by {\modelname} and {\modelname} w/o Meta Learning on the unlabeled data in all 10 iterations. Both models are evaluated with 10\% labeled data and 50\% unlabeled data on the SemEval dataset. Note that following \citet{li2020exploit}, we use stratified sampling to obtain unlabeled data for each iteration, so they all share the same gold label distribution. From Figure \ref{fig:label_distribution}, we observe that with Meta Learning, the pseudo label distribution is closer to the gold label distribution, with less drift, and thus contributes to the robust iterative self-training schema.

\noindent\textbf{Effectiveness of Incremental Self-Training}

During each iteration of the Incremental Self-Training, the Pseudo Label Selection module selects the top 90\% high-confidence pseudo labels which contain less noise and abandon the rest. From Figure \ref{fig:meta-learning}, we could observe that on both SemEval and TACRED dataset, the selection mechanism achieves nearly 5\% - 10\% F1 performance improvement compared to the full, unfiltered pseudo labels. The improvement is higher on the TACRED dataset, which has a more skewed label distribution.

\begin{figure}[t!]
    \centering
    \includegraphics[width=0.49\linewidth]{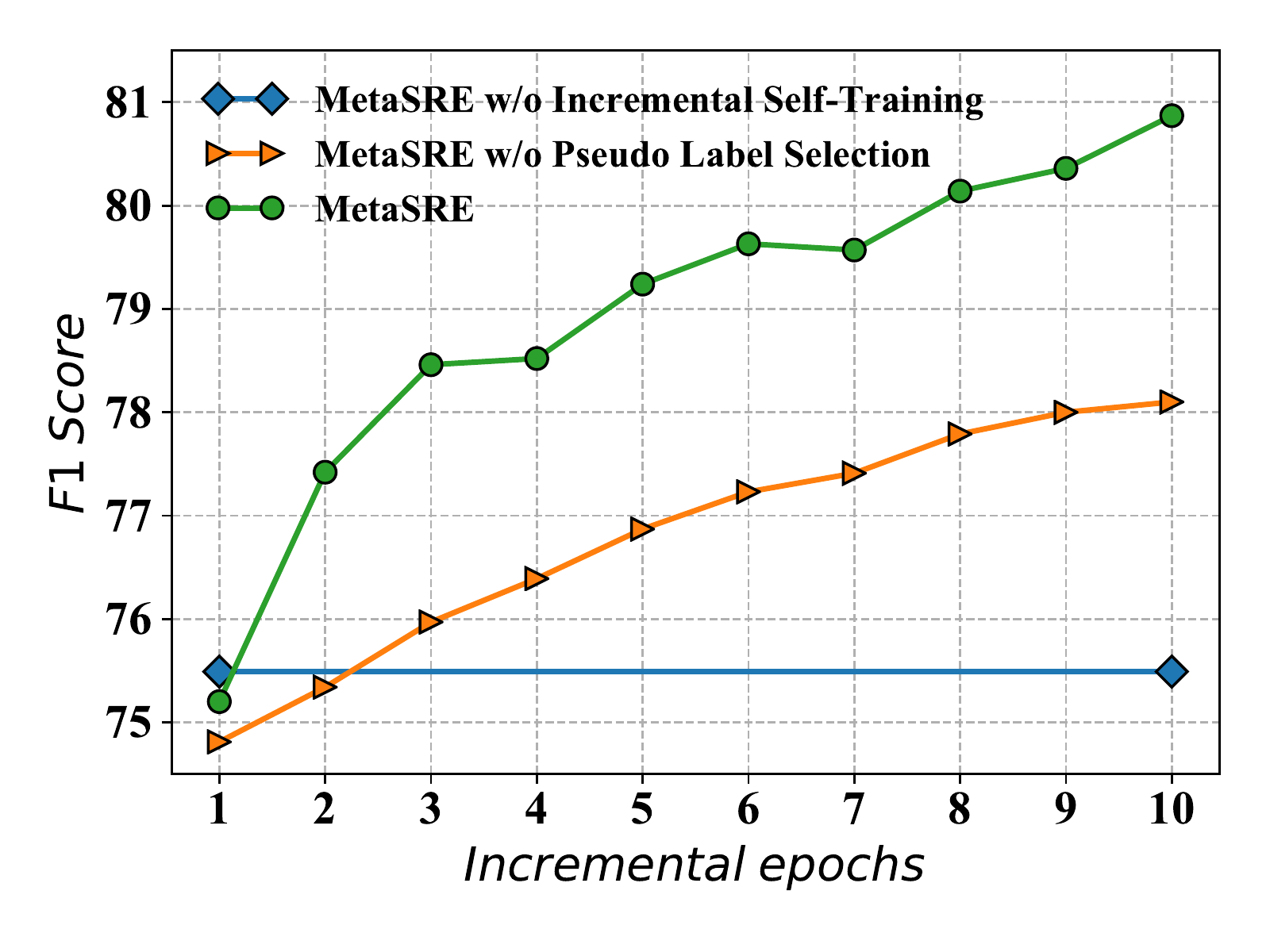} 
    \includegraphics[width=0.49\linewidth]{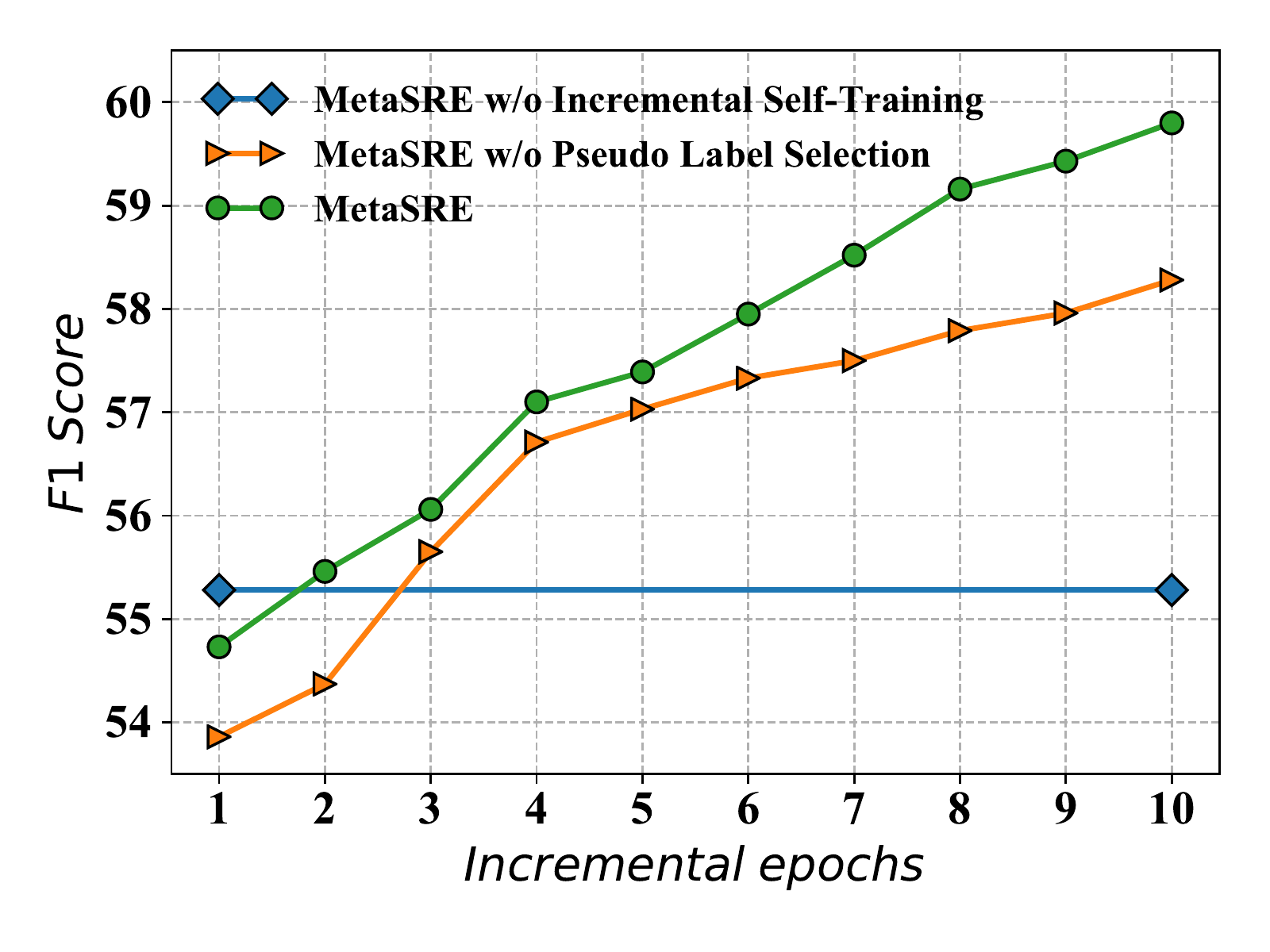}
    \caption{Effectiveness of Incremental Self-Training module on SemEval (left) and TACRED (right).}
    \label{fig:Incremental}
\vspace{-0.1in}
\end{figure}

To demonstrate that high-confidence pseudo labels selected by Pseudo Label Selection could be helpful for robust self-training, we compare Incremental Self-Training that uses 10\% unlabeled data with high-confidence pseudo labels in each iteration, with the same ratio unfiltered pseudo labels on unlabeled data. We sample the SemEval with 10\% and the TACRED with 15\% labeled data, both with 50\% unlabeled data. From Figure \ref{fig:Incremental}, when comparing with {\modelname} w/o Pseudo Label Selection ({\color{brown}brown}), the Incremental Self-Training handles noises in pseudo label selection \& exploitation more effectively, leading to robust and consistently better performance.

To explore the effectiveness of the incremental scheme, we report the F1 on {\modelname} w/o Incremental Self-Training ({\color{blue}blue}), which adds all the pseudo labels to the labeled data at once to optimize {\f}. From Figure \ref{fig:Incremental}, {\modelname} ({\color{green}green}) could obtain more competitive results compared with blue line, which is mainly because {\modelname} leverages the most up-to-date {\g} to generate higher quality pseudo labels and incrementally improves the classification performance.

\begin{table}[]
\centering
\resizebox{.8\linewidth}{!}{
\begin{tabular}{lccccccc}
\toprule
 \diagbox{Dataset}{${Z\%}$}    & 60    & 70    & 80    & 90    & 100   \\
\midrule
 SemEval & 74.89 & 75.26 & 78.36 & \textbf{80.09} & 77.28 \\
 TACRED  & 53.38 & 55.12 & 56.60 & \textbf{56.95} & 55.56 \\

\bottomrule
\end{tabular}}
\caption{$F_{1}$ Performance with different ${Z\%}$ on 10\% labeled data of SemEval and TACRED.}
\label{tab:TopZ}
\vspace{-0.1in}
\end{table}

\noindent\textbf{Hyperparameter Analysis}

We study the ratio of top ${Z\%}$ in Pseudo Label Selection module. We sample ${Z\%}$ from 60\% to 100\% and report the $F_{1}$ Performance of SemEval and TACRED with the same 10\% labeled data. Note that our hyperparameter ${Z\%}$ is decided based on the validation set and we report the ${Z\%}$ analysis on the test set here for better demonstration. From Table \ref{tab:TopZ}, the fluctuation results indicate that both quality and coverage of pseudo labels will impact performance. Using a high ${Z\%}$ will introduce low-quality pseudo labels that are noisy-prone, causing the gradual drift problem. Low ${Z\%}$ will cause the low coverage of pseudo labels on some relations, affecting the recall.

\section{Related Work}
Relation extraction focuses on predicting the relation between two entities in a sentence. Recent literature leverage deep neural networks to extract features about two entities from sentences, and classify them into specific relations. Relation extraction methods are often formulated in a supervised setting \citep{zeng2015distant,DBLP:conf/ijcai/GuoN0C20,DBLP:conf/acl/NanGSL20} and require manual annotations on large amounts of corpora, which is labor-intensive to obtain.

Semi-supervised learning methods have received attention recently, since these methods require fewer labeled data and generate pseudo labels by re-training model and improve the performance iteratively. 
Two major categories of semi-supervised learning methods are related to our problem.
One major category is the self-ensembling method \citep{french2017self,xu2019self}, which is based on the assumption that pseudo label distributions should remain unchanged even if the model parameters and instances have small perturbations. In this case, various models based on consistent data could be used to improve the performance by co-training each other \citep{tarvainen2017mean}. However, this method relies heavily on the quality and quantity of labeled data. In our task, the improvement of the model by this method is limited.

Another category is the self-training method, the work proposed by \citet{rosenberg2005semi} incrementally generates pseudo labels from unlabeled data, and uses these pseudo labels to enhance the classification ability of the model. However, this method often needs to endure semantic drift \citep{curran2007minimising,zhang2016understanding,hu2020selfore,li2020exploit}, which means the noise generated by pseudo labels in each iteration will be continuously strengthened, causing the model to deviate from the global minimum point. \citet{li2019learning} tries to fine-tune the model on the labeled data in each iteration to avoid noise and construct a prototypical network to exploit pseudo labels. However, the model is still influenced by the noises with the unchanged pseudo labels and spreads these noises to the generated pseudo labels. In our work, we adopt a meta learner network to prevent the model from drifting due to label noise and enables robust iterative self-training.

\section{Conclusion}
In this paper, we propose a semi-supervised learning model {\modelname} for relation extraction. Different from conventional semi-supervised models which directly adopt the classification network to classify unlabeled data into pseudo labels, our model proposes a novel meta-learning-based self-training network to reduce the noises contained in pseudo labels and avoid the model gradual drift. Comparing with using pseudo labels directly, our model leverages pseudo label selection and exploitation scheme to further select high-confidence pseudo labels with low noises. Experiments on two popular benchmarks show the effectiveness and consistent improvements over baselines.
\section*{Acknowledgments}
We thank the reviewers for their valuable comments. The work was supported by the National Key Research and Development Program of China (No. 2019YFB1704003), the National Nature Science Foundation of China (No. 71690231 and No. 62021002), NSF under grants III-1763325, III-1909323, III-2106758, SaTC-1930941, Tsinghua BNRist and Beijing Key Laboratory of Industrial Bigdata System and Application.
\balance
\bibliography{anthology,custom}

\begin{thebibliography}{25}
\expandafter\ifx\csname natexlab\endcsname\relax\def\natexlab#1{#1}\fi

\bibitem[{Curran et~al.(2007)Curran, Murphy, and Scholz}]{curran2007minimising}
James~R Curran, Tara Murphy, and Bernhard Scholz. 2007.
\newblock Minimising semantic drift with mutual exclusion bootstrapping.
\newblock In \emph{Proc. of PACL}, volume~6, pages 172--180. Bali.

\bibitem[{Devlin et~al.(2019)Devlin, Chang, Lee, and
  Toutanova}]{devlin2019bert}
Jacob Devlin, Ming-Wei Chang, Kenton Lee, and Kristina Toutanova. 2019.
\newblock Bert: Pre-training of deep bidirectional transformers for language
  understanding.
\newblock In \emph{Proc. of NAACL-HLT}, pages 4171--4186.

\bibitem[{Finn et~al.(2017)Finn, Abbeel, and Levine}]{finn2017model}
Chelsea Finn, Pieter Abbeel, and Sergey Levine. 2017.
\newblock Model-agnostic meta-learning for fast adaptation of deep networks.
\newblock In \emph{Proc. of ICML}, pages 1126--1135. JMLR. org.

\bibitem[{French et~al.(2017)French, Mackiewicz, and Fisher}]{french2017self}
Geoffrey French, Michal Mackiewicz, and Mark Fisher. 2017.
\newblock Self-ensembling for visual domain adaptation.
\newblock \emph{arXiv preprint arXiv:1706.05208}.

\bibitem[{Guo et~al.(2020)Guo, Nan, Lu, and Cohen}]{DBLP:conf/ijcai/GuoN0C20}
Zhijiang Guo, Guoshun Nan, Wei Lu, and Shay~B. Cohen. 2020.
\newblock Learning latent forests for medical relation extraction.
\newblock In \emph{Proc. of IJCAI}, pages 3651--3657. ijcai.org.

\bibitem[{Hendrickx et~al.(2010)Hendrickx, Kim, Kozareva, Nakov, S{\'e}aghdha,
  Pad{\'o}, Pennacchiotti, Romano, and Szpakowicz}]{hendrickx2010semeval}
Iris Hendrickx, Su~Nam Kim, Zornitsa Kozareva, Preslav Nakov, Diarmuid~O
  S{\'e}aghdha, Sebastian Pad{\'o}, Marco Pennacchiotti, Lorenza Romano, and
  Stan Szpakowicz. 2010.
\newblock Semeval-2010 task 8: Multi-way classification of semantic relations
  between pairs of nominals.
\newblock In \emph{Proc. of the 5th International Workshop on Semantic
  Evaluation}, pages 33--38.

\bibitem[{Hochreiter and Schmidhuber(1997)}]{hochreiter1997long}
Sepp Hochreiter and J{\"u}rgen Schmidhuber. 1997.
\newblock Long short-term memory.
\newblock \emph{Neural computation}, 9(8):1735--1780.

\bibitem[{Hu et~al.(2020)Hu, Wen, Xu, Zhang, and Philip}]{hu2020selfore}
Xuming Hu, Lijie Wen, Yusong Xu, Chenwei Zhang, and S~Yu Philip. 2020.
\newblock Selfore: Self-supervised relational feature learning for open
  relation extraction.
\newblock In \emph{Proc. of EMNLP}, pages 3673--3682.

\bibitem[{Li and Qian(2020)}]{li2020exploit}
Wanli Li and Tieyun Qian. 2020.
\newblock Exploit multiple reference graphs for semi-supervised relation
  extraction.
\newblock \emph{arXiv preprint arXiv:2010.11383}.

\bibitem[{Li et~al.(2019)Li, Sun, Liu, Zhou, Zheng, Chua, and
  Schiele}]{li2019learning}
Xinzhe Li, Qianru Sun, Yaoyao Liu, Qin Zhou, Shibao Zheng, Tat-Seng Chua, and
  Bernt Schiele. 2019.
\newblock Learning to self-train for semi-supervised few-shot classification.
\newblock In \emph{NeurIPS}, pages 10276--10286.

\bibitem[{Lin et~al.(2019)Lin, Yan, Qu, and Ren}]{lin2019learning}
Hongtao Lin, Jun Yan, Meng Qu, and Xiang Ren. 2019.
\newblock Learning dual retrieval module for semi-supervised relation
  extraction.
\newblock In \emph{Proc. of WWW}, pages 1073--1083.

\bibitem[{Liu et~al.(2019)Liu, Davison, and Johns}]{liu2019self}
Shikun Liu, Andrew Davison, and Edward Johns. 2019.
\newblock Self-supervised generalisation with meta auxiliary learning.
\newblock In \emph{NeurIPS}, pages 1677--1687.

\bibitem[{Mintz et~al.(2009)Mintz, Bills, Snow, and
  Jurafsky}]{mintz2009distant}
Mike Mintz, Steven Bills, Rion Snow, and Dan Jurafsky. 2009.
\newblock Distant supervision for relation extraction without labeled data.
\newblock In \emph{Proc. of ACL}, pages 1003--1011.

\bibitem[{Miyato et~al.(2018)Miyato, Maeda, Koyama, and
  Ishii}]{miyato2018virtual}
Takeru Miyato, Shin-ichi Maeda, Masanori Koyama, and Shin Ishii. 2018.
\newblock Virtual adversarial training: a regularization method for supervised
  and semi-supervised learning.
\newblock \emph{IEEE TPAMI}, 41(8):1979--1993.

\bibitem[{Nan et~al.(2020)Nan, Guo, Sekulic, and Lu}]{DBLP:conf/acl/NanGSL20}
Guoshun Nan, Zhijiang Guo, Ivan Sekulic, and Wei Lu. 2020.
\newblock Reasoning with latent structure refinement for document-level
  relation extraction.
\newblock In \emph{Proc. of ACL}, pages 1546--1557.

\bibitem[{Rosenberg et~al.(2005)Rosenberg, Hebert, and
  Schneiderman}]{rosenberg2005semi}
Chuck Rosenberg, Martial Hebert, and Henry Schneiderman. 2005.
\newblock Semi-supervised self-training of object detection models.
\newblock \emph{WACV/MOTION}, 2.

\bibitem[{Soares et~al.(2019)Soares, FitzGerald, Ling, and
  Kwiatkowski}]{soares2019matching}
Livio~Baldini Soares, Nicholas FitzGerald, Jeffrey Ling, and Tom Kwiatkowski.
  2019.
\newblock Matching the blanks: Distributional similarity for relation learning.
\newblock \emph{arXiv preprint arXiv:1906.03158}.

\bibitem[{Tarvainen and Valpola(2017)}]{tarvainen2017mean}
Antti Tarvainen and Harri Valpola. 2017.
\newblock Mean teachers are better role models: Weight-averaged consistency
  targets improve semi-supervised deep learning results.
\newblock In \emph{NeurIPS}, pages 1195--1204.

\bibitem[{Xu et~al.(2019)Xu, Du, Zhang, Zhang, Wang, and Zhang}]{xu2019self}
Yonghao Xu, Bo~Du, Lefei Zhang, Qian Zhang, Guoli Wang, and Liangpei Zhang.
  2019.
\newblock Self-ensembling attention networks: Addressing domain shift for
  semantic segmentation.
\newblock In \emph{Proc. of AAAI}, volume~33, pages 5581--5588.

\bibitem[{Zeng et~al.(2015)Zeng, Liu, Chen, and Zhao}]{zeng2015distant}
Daojian Zeng, Kang Liu, Yubo Chen, and Jun Zhao. 2015.
\newblock Distant supervision for relation extraction via piecewise
  convolutional neural networks.
\newblock In \emph{Proc. of EMNLP}, pages 1753--1762.

\bibitem[{Zeng et~al.(2016)Zeng, Lin, Liu, and Sun}]{zeng2016incorporating}
Wenyuan Zeng, Yankai Lin, Zhiyuan Liu, and Maosong Sun. 2016.
\newblock Incorporating relation paths in neural relation extraction.
\newblock \emph{arXiv preprint arXiv:1609.07479}.

\bibitem[{Zhang et~al.(2016)Zhang, Bengio, Hardt, Recht, and
  Vinyals}]{zhang2016understanding}
Chiyuan Zhang, Samy Bengio, Moritz Hardt, Benjamin Recht, and Oriol Vinyals.
  2016.
\newblock Understanding deep learning requires rethinking generalization.
\newblock \emph{arXiv preprint arXiv:1611.03530}.

\bibitem[{Zhang and Wang(2015)}]{zhang2015relation}
Dongxu Zhang and Dong Wang. 2015.
\newblock Relation classification via recurrent neural network.
\newblock \emph{arXiv preprint arXiv:1508.01006}.

\bibitem[{Zhang et~al.(2018)Zhang, Tang, and Jia}]{zhang2018fine}
Yabin Zhang, Hui Tang, and Kui Jia. 2018.
\newblock Fine-grained visual categorization using meta-learning optimization
  with sample selection of auxiliary data.
\newblock In \emph{Proc. of ECCV}, pages 233--248.

\bibitem[{Zhang et~al.(2017)Zhang, Zhong, Chen, Angeli, and
  Manning}]{zhang2017position}
Yuhao Zhang, Victor Zhong, Danqi Chen, Gabor Angeli, and Christopher~D Manning.
  2017.
\newblock Position-aware attention and supervised data improve slot filling.
\newblock In \emph{Proc. of EMNLP}, pages 35--45.

\end{thebibliography}
\bibliographystyle{acl_natbib}

\end{document}